\documentclass[conference]{IEEEtran}
\IEEEoverridecommandlockouts
\usepackage{amsmath,amssymb,amsfonts}
\usepackage{algorithmic}
\usepackage{graphicx}
\usepackage{textcomp}
\usepackage{xcolor}
\def\BibTeX{{\rm B\kern-.05em{\sc i\kern-.025em b}\kern-.08em
    T\kern-.1667em\lower.7ex\hbox{E}\kern-.125emX}}

\usepackage[
    backend=biber,
    style=ieee,
    sorting=none,
    natbib=true,
    url=true,
    doi=true,
    eprint=false,
    maxnames=5,
    maxcitenames=3,
    autocite=superscript
]{biblatex}
\graphicspath{ {./images/} }
    
\begin{document}

\title{CISRU: a robotics software suite to enable complex rover-rover and astronaut-rover interaction.\\
\thanks{The CISRU project is partially funded by ESA under grant agreement No. 4000135391/21/NL/GLC/zk}
\thanks{Corresponding authors: Alba Guerra, aguerra@gmv.com; Cristina Luna, cluna@gmv.com}
}

 \author{%
        \IEEEauthorblockN{
            Silvia Romero-Azpitarte%
            \IEEEauthorrefmark{1},
            Alba Guerra%
            \IEEEauthorrefmark{1},
            Mercedes Alonso%
            \IEEEauthorrefmark{1},
            Marina L. Seoane%
            \IEEEauthorrefmark{1},\\
            Daniel Olayo%
            \IEEEauthorrefmark{1},
            Almudena Moreno%
            \IEEEauthorrefmark{1},
            Pablo Castellanos%
            \IEEEauthorrefmark{1},
            Cristina Luna%
            \IEEEauthorrefmark{1},
            Gianfranco Visentin%
            \IEEEauthorrefmark{2}
        }
        \IEEEauthorblockA{
            \IEEEauthorrefmark{1}
            GMV Aerospace and Defence SAU, 
            Tres Cantos, Spain}

        \IEEEauthorblockA{%
            \IEEEauthorrefmark{2}
            ESTEC, ESA, 
            Noordwijk, the Netherlands
        }
    }


\maketitle
\thispagestyle{plain}
\pagestyle{plain}

\begin{abstract}
The CISRU project has focused on the development of a software suite for planetary (and terrestrial) robotics, fully abstracted from the robotic platform and enabling interaction between rovers and astronauts in complex tasks and non-structured scenarios. To achieve this, a high level of autonomy is required, powered by AI and multi-agent autonomous planning systems inherited from ERGO/ADE \citep{CISRU2022} and the PERASPERA program. This communication presents the system developed in CISRU, focusing on the modules of AI-based perception and the interaction between astronauts and robots.
\end{abstract}


\section{Introduction}

Space exploration, particularly the long-term habitation of planetary surfaces, requires significant technological advances, with a strong focus on collaboration between robots and astronauts where the modularity and autonomy of space robots will stand out, allowing them to perform different tasks \citep{luna2023modularity, Luna2023ELS}. Efficient utilisation of space resources during the establishment of extraterrestrial settlements is also of paramount importance.

In the AI-enabled robotics SW suite for autonomous Collaborative ISRU (CISRU) project \citep{romero2023enabling}, we have developed a comprehensive software suite that facilitates our understanding, navigation, and interaction with the environment and its agents, including robots and astronauts. The CISRU suite comprises five main modules, each designed to fulfil specific objectives.

The first module centres on multi-agent autonomy components and enables seamless communication among different agents and mission control. The second module is dedicated to perception and incorporates AI algorithms to enhance environmental awareness. This encompasses environment segmentation, object and agent pose estimation, obstacle detection, as well as damage and emergency situation identification.

The third module provides essential components for safe navigation, encompassing obstacle avoidance, social navigation with astronauts, and collaboration among diverse robots. The fourth module focuses on manipulation functions, which play a crucial role in In-Situ Resource Utilisation (ISRU) scenarios. CISRU integrates multi-tool manipulation functions, a novel tool-changer design, and a variety of objects, empowering agents to autonomously undertake a wide range of tasks.

The fifth module oversees the cooperative behaviour of all modules, integrating astronaut command and Mixed Reality interfaces. It also includes map fusion of different agents, task supervision, as well as emergency and error control. To validate the suite's performance, an astronaut-rover interaction dataset within a dedicated planetary environment have been generated. Additionally, extensive testing has been conducted in the GMV SPoT analog environment and various simulators.

The test results exemplify the advantages of the E4 level of autonomy (following the levels defined by ECSS standards), enabling a high level of abstraction and showcasing the capabilities of AI in space systems. This level of autonomy, in conjunction with collaboration between astronauts and robots, is pivotal for the successful construction of structures and the accomplishment of mission-specific tasks. This paper presents the development of the CISRU suite, the preparation of field tests, and the analysis of results, highlighting the potential of this AI-powered comprehensive suite and emphasising the significance of high autonomy and collaboration in paving the way for future planetary exploration missions.

\section{CISRU structure}
In this section, we describe the overall architecture of the developed software suite. The suite, as mentioned above, is divided into 5 different modules, that are designed to work simultaneously and to exchange different data to obtain a successful result. The perception components are described in Section III given the importance of such development.

The different interfaces of the modules were developed accordingly, and, in some cases, they make use of previous technologies such as ROS2 interface messages or ERGO Agent messages. The system is structured in such a way that the control centre is able to communicate with all the robot Agents and then each Agent is able to convert the high level commands into the different subsystem medium level commands like moving to a specific position as part of locomotion subsystem. Meanwhile both robots can make use of the Perception and Navigation components.

\subsection{Multi-agent component}
Based on the ERGO architecture, for the communication with the different subsystem reactors, the agent controller uses a specific interface based on goals and observations.
In a multi-agent system, two or more agents interact with each other to accomplish more complex tasks. 
Using a specific reactor, the MAS (Multi-Agent Synchronisation) reactor, the different agent instances can send and receive goals and observations from the other agents. 
The agents can work in different autonomy levels. This autonomy levels follow the ECSS standard and determine what telecommands are processed and the level of autonomous decision taking enabled. The autonomy level acquired for CISRU is E4 autonomy level. Only E4 goals are accepted in this level. E4 goals are high level goals that the system decomposes, plans and schedules using on-board autonomy and decision-making capabilities (typically with a planner on-board). If the execution deviates from the expected, the system can adapt the plan accordingly.
This Multi-Agent is also in charge of the activities synchronisation between the different vehicles or agents, specifically for this project two rovers and one astronaut. This component is required for improving the collaboration between the different agents and being able to adapt the plan depending on the status of the different agents.

\subsection{Guidance, navigation and control components}

The Guidance, navigation and control (GNC) of the rovers is necessary in order to be able to move the vehicles autonomously. The navigation component is based on Visual SLAM (Simultaneous Localisation and Mapping) using information from the stereo camera, inertial sensor and wheel odometry. The input from the stereo camera is processed by a perception component, as detailed in the perception section, to obtain a filtered point cloud and used as SLAM input to improve the performance of the algorithm.

\begin{figure}[h!]
    \centering
    \includegraphics[width=\columnwidth]{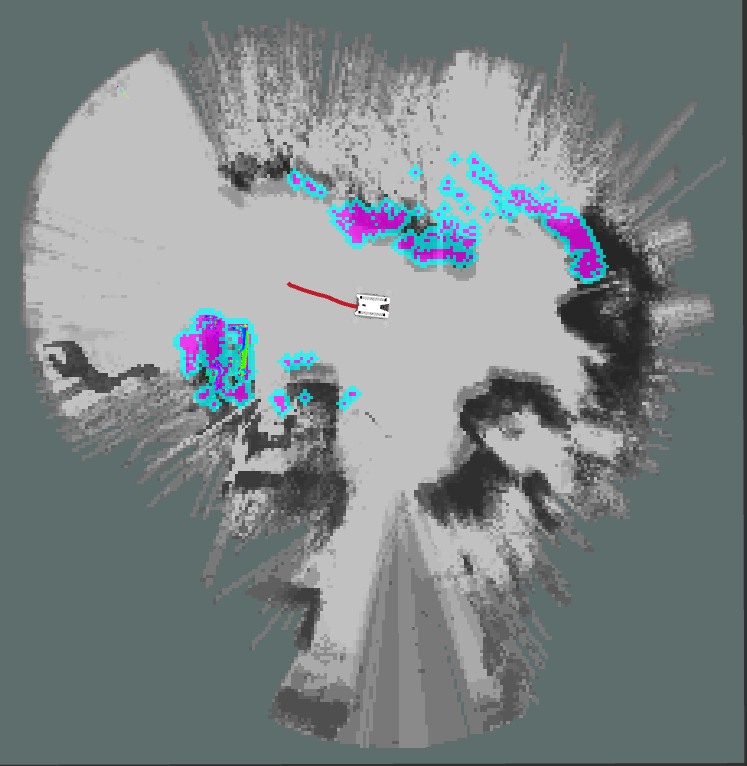}
    \caption{Fast Marching Square algorithm}
    \label{fig:fm2}
\end{figure}

The guidance and control is based on ROS2 packages like NAV2 which is a software library to obtain a trajectory plan and control robotic vehicles \citep{macenski2020marathon2}. For the path planning, a NAV2 plugin had been developed based on Fast Marching Square algorithm \citep{FMS} and integrated with the control.

\subsection{Manipulation components}
The manipulation components are required to obtain samples simulating a exploration mission where the rovers should map a new zone and take samples from the interesting regions. Also the manipulator should be useful for astronaut collaboration. Due to these requirements, the main rover has a ROBOTIS Manipulator-H robotic arm with a specific tool-changer on the end-effector to perform the different tasks.

The control of the robotic arm is based on MoveIt2 from PickNik which is a ROS2 package prepared to do path planning and control different robotic arms. The first component of the manipulation is able to communicate with MoveIt2 to command different positions to the robotic arm. The second component is the tool-changer which is in charge of all the procedure to assembling and disassembling the different tools by communicating with the other components and the management of the tool status. The last component is the sample collection, which is also capable to communicate with the other components to assembly a shovel tool, collect a sample and storage it.

\subsection{Cooperative behaviour components}
As the name of CISRU project indicates, one of the main objectives to improve the robotics and astronaut collaboration, so the cooperative behaviour components are the ones necessary for this objective.

The robot collaboration requires to share information and knowing the position of the other vehicles. For this reason, the environment detection is also able to detect the rovers which improves the collaboration by avoiding possible path collision between both vehicles. The second component, to improve this collaboration is the mapping fusion which is necessary to know the position of the both robots based on same map. 

\begin{figure}[h!]
    \centering
    \includegraphics[width=\columnwidth]{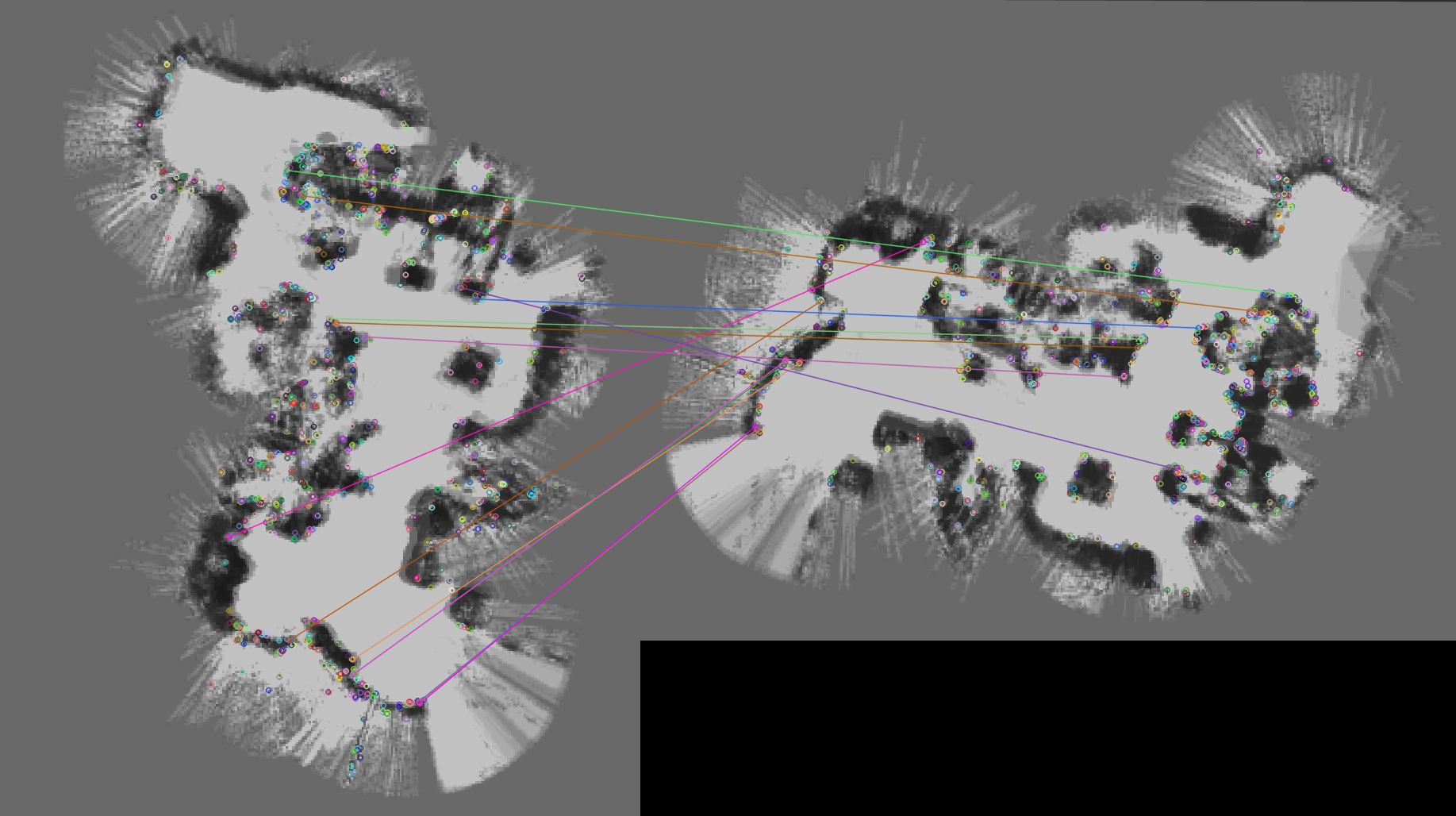}
    \caption{Map fusion}
    \label{fig:map fusion}
\end{figure}

In the design of the Human-Robot Interaction (HRI) component, a shared agency between astronauts and robots has been employed as the ontological foundation. This allows for commanding both the astronaut and the different rovers, while they share the same information and are capable of supervising each other.

To enable this, the astronaut is equipped with technical devices that facilitate communication through various methods with the mission control centre and the different Robotics Working Agents (RWAs). In this case, a device based on Microsoft Hololens 2 has been developed, which allows the astronaut to visualise real-time information and send commands to and receive commands from the RWAs and the mission control centre. Additionally, a console integrated into the forearm of the spacesuit serves as a monitor, facilitating the necessary levels of communication, primarily through manual interaction with the screen.

The mixed reality (MR) device enables the astronaut to visualise and interpret information, as well as issue commands using gaze and voice - although gestures can also be utilised. This capability is particularly valuable during extravehicular activities (EVAs), as it allows the astronaut to manage tasks without having to release the tools or equipment they are using at that moment.

In this system, as previously mentioned, the rovers are capable of monitoring the astronaut and detecting any emergencies, such as if the astronaut falls, deviates from the expected location, or loses communication. This represents a significant improvement in risk management for each mission.

\section{Perception component}

The perception component in CISRU was one of the most important developments. These components oversee the vision and perception of the robots. They include Human-Machine interaction detection, equipment anomaly detection, and Human emergency situations, as well as semantic segmentation for the correct identification of obstacles for navigation.

These functions were implemented as Neural Networks. All models but semantic segmentation one,  are low resource consumption and run in a MyriadX VPU, a computing unit that is being tested in space environments as a possibility to process AI models in extreme conditions \citep{ramaswami2022single}. The semantic segmentation model, given that it is bigger and more exhaustive, is prepared to run on a GPU or FPGA (Zynq UltraScale+ MPSoC is the current chosen hardware, which is also being tested for spatial environments \citep{hiemstra2017single}).

The particular architecture used is for all models but semantic segmentation is Mobilenet-SSD \citep{howard2017mobilenets}, but trained for different targets. This architecture is dedicated to image detection, and it uses Single Shot Detection as one of the key features for its fast and precise output \citep{liu2016ssd}. It was selected because of the input data received (a continuous video) that permitted us to have a precise result that could be corrected over time thanks to the fast predictions of the model. 

On the following subsections, we explain each of the functionalities and how the inside mechanism works, considering the input image and the output detections. All output detections of all models are then integrated and used in ROS2 components, which were described briefly in the CISRU structure section.

\subsection{Astronaut interaction detection and emergency situations}
These two functionalities were seamlessly integrated into a unified system by training a single MobileNet-SSD model instance and incorporating a tracking algorithm atop the output layer. This innovative approach enabled instance identification across successive images, proving to be an efficient solution for the CISRU scenario.

The tracking algorithm itself, based on the Kalman filter, played a pivotal role in maintaining lightweight computing demands \citep{henriques2014high}. Although it may not offer the highest level of accuracy, it successfully tracked objects detected in the image stream. While there was a possibility of reassigning an instance if a very similar object of the same class appeared in the image, this scenario was unlikely given the sparse presence of astronauts and robots in the environment. Additionally, given the hardware constraints of the mission, this tracking system proved to be highly effective, allowing for the simultaneous tracking of up to 60 objects.

The model's training process leveraged a dataset containing detection labels for astronauts, rovers, rocks, and solar panels. The input to the system comprised two images: one for depth (derived from merging stereo images) and one for RGB information. The RGB image was processed by the network, generating a vector of bounding boxes representing the instances found in each image.

To associate spatial coordinates with these regions of interest (ROI), a spatial location calculator was employed. It utilised depth information from the input depth map to calculate the average depth values within the ROIs and removed those falling out of the specified range. These bounding boxes, along with their associated depth information, were then fed into the tracking algorithm, which determined the identity of each object. If an object had not been previously identified, the algorithm incremented its ID, ensuring consistent tracking across subsequent detections.

\begin{figure}[h!]
    \centering
    \includegraphics[width=\columnwidth]{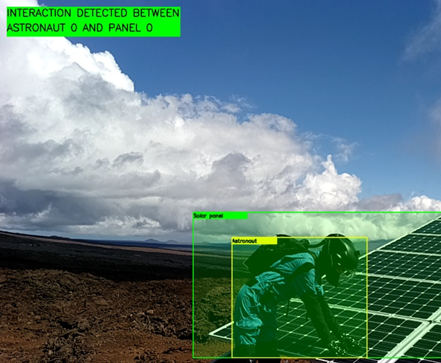}
    \caption{Human interaction detected between astronaut and a solar panel at HI-SEAS}
    \label{fig:hmid}
\end{figure}
\begin{figure}
    \centering
    \includegraphics[width=\columnwidth]{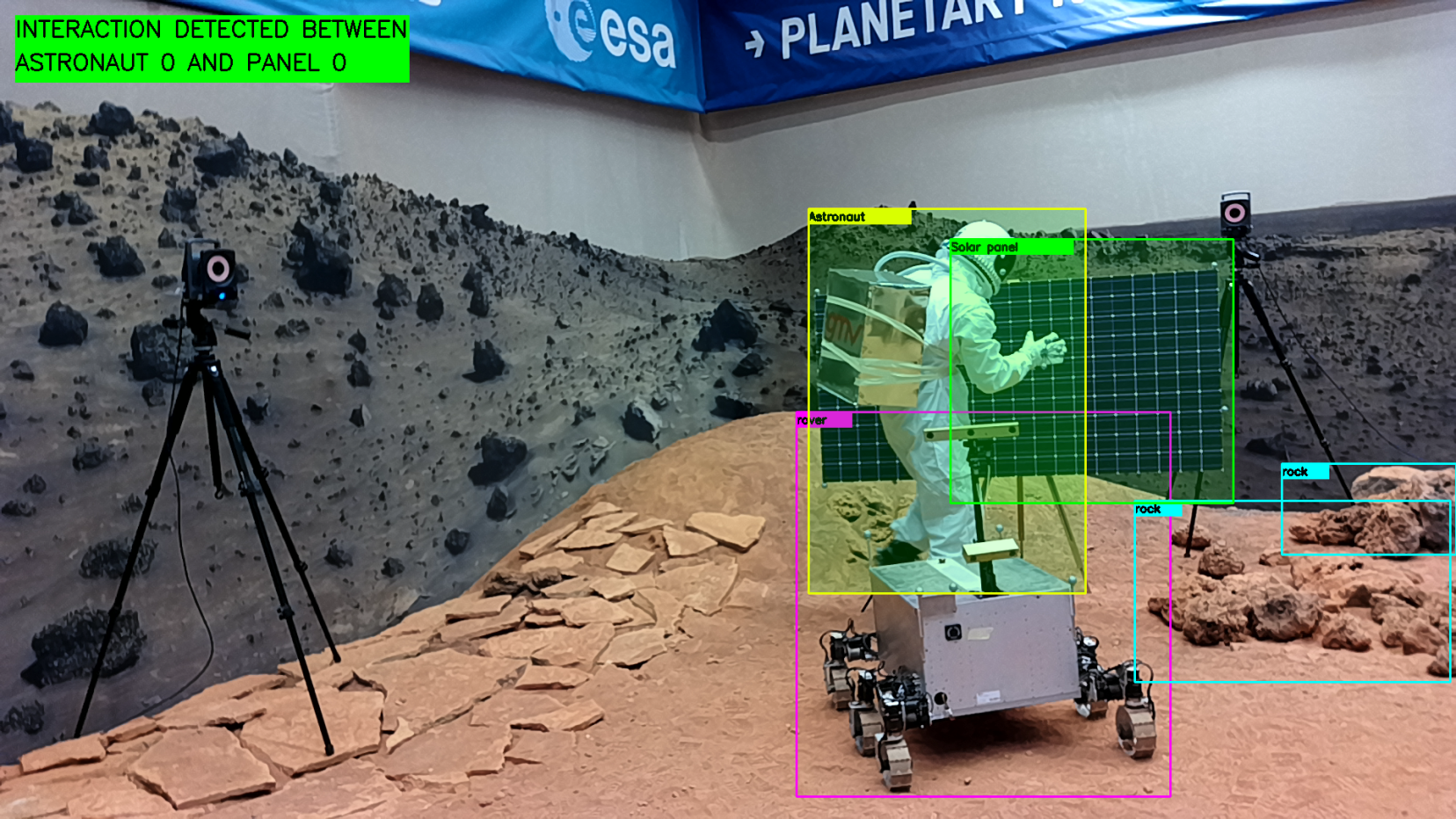}
    \caption{Human interaction detected between astronaut and a solar panel at ESTEC}
    \label{fig:hmid_estec}
\end{figure}

Building on these fundamental calculations, a higher-level function specialised in recognising instances of astronauts within predefined interactive distances from rovers or solar panels, identifying it as Human-Machine interaction. Additionally, the system was designed to detect emergencies, such as astronauts being in close proximity to dangerous objects like rocks. The definitions of dangerous labels were configurable, and this configuration layer did not interfere with the lower-level software, showcasing the system's modular design.

Emergencies were also detected in the event of an astronaut falling. This anomaly was identified by analysing the tracked instance's pose and position, with sudden changes in bounding box disposition serving as a key indicator \citep{yao2017new}. Different values for the gravitational constant were taken into account to accommodate variations in the extraterrestrial environment.

In response to these emergency situations, the system promptly notified the multi-agent system, ensuring that appropriate measures were taken to safeguard the mission's success and the well-being of astronauts and robots alike. This sophisticated integration of object detection, tracking, and emergency response mechanisms showcased the robustness and adaptability of the CISRU scenario's monitoring and interaction system.

\subsection{Equipment anomaly detection}
This functionality serves a highly specialised purpose, specifically tailored for the CISRU scenario where the focus is on analysing solar panels distributed within a test field. In such extraterrestrial environments, solar panels play a pivotal role as a primary source of energy alongside nuclear power sources. The key challenges for solar panels in these spatial settings primarily revolve around two main issues: the risk of cracking due to impacts, which can be mitigated by the use of protective glasses, and the concern of overheating \citep{plis2021effect},  \citep{kawakita2009discharge}. These concerns formed the basis for labelling the dataset of solar panels.

The dataset was meticulously curated for image detection, employing data augmentation techniques on various mock-ups of solar panels under diverse illumination conditions and incorporating different patches of cracks and burnouts to ensure a comprehensive representation of real-world scenarios.

To facilitate the analysis, a MobileNet-SSD architecture was selected as the model of choice. This model had been trained using a dataset specifically focused on identifying cracks on solar panels. Whenever a rover was directed to autonomously inspect a rack of solar panels, the model was deployed on the MyriadX VPU. The command to the rover included crucial parameters, notably the number of solar panels comprising a rack. Each individual solar panel was uniquely identified using April tags.

The rover's inspection process entailed a systematic examination of the rack. For every April tag detected, the model would evaluate whether it could view the entire solar panel. If it could, and the panel was intact, it would be recorded as 'good.' In contrast, if the model couldn't observe the entire panel, it was marked as 'spotted.' Additionally, if any part of the panel that was visible was found to be cracked, it would be promptly categorised as 'cracked.'

\begin{figure}[h!]
    \centering
    \includegraphics[width=\columnwidth]{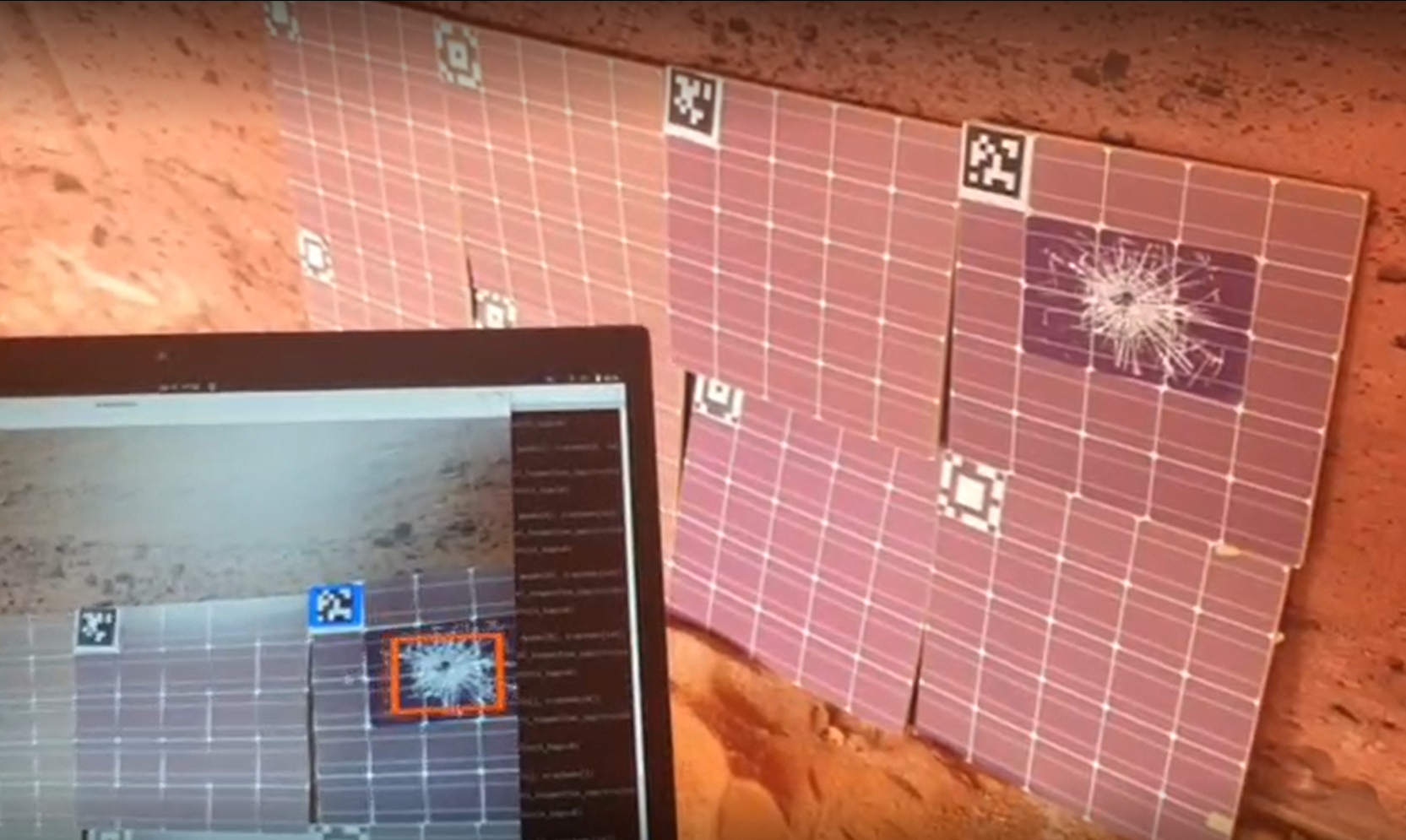}
    \caption{Semantic segmentation module}
    \label{fig:carckdets}
\end{figure}

This methodology exhibited a high level of accuracy, which proved essential given the challenging terrain through which the rover navigated. The uneven terrain often caused the camera to be off-centre, making the distinctions between 'good,' 'spotted,' and 'cracked' panels, as well as the transitions between these states, invaluable in ensuring the efficiency and reliability of the inspection process.

The training of this model followed a straightforward approach, as it was essentially an adaptation of the previous model, albeit without the additional layers. This simplicity in model development and the rigor in dataset preparation contributed to the success of the solar panel inspection system within the CISRU scenario.

\subsection{Semantic segmentation}
The functionality of semantic segmentation represents a critical component of the system, integral to the mapping and navigation module. Its significance arises from the unique challenges posed by the choice of navigation sensor: a Stereo-RGB camera. Unlike LIDAR and similar systems, stereo cameras are known for their limitations in generating accurate depth maps  \citep{campos2021orb}, adding complexity due to the absence of a precise ground truth in the context of uneven terrains found on celestial bodies like Mars, the Moon, or the environments used for testing. These terrains, when mapped using a stereo camera, often produce noisy point clouds or depth maps, rendering them unsuitable for safe navigation by robotic autonomous systems.

To address this issue, a semantic segmentation model was introduced to interpret the images and determine which volumes, as detected in the depth image, pose actual navigation hazards. Semantic segmentation, unlike other neural network approaches such as object detection, involves recreating and reinterpreting the entire input image to assign each pixel a label corresponding to the object it belongs to. However, this computational process is resource-intensive, demanding more powerful hardware. Consequently, the system transitioned to more robust options, specifically the Zynq UltraScale+ MPSoC and Jetson Orin AGX, both of which were tested to benchmark the model's performance in real-time navigation scenarios.

The dataset preparation for this functionality is an extensive and meticulous process, as elaborated in the AI dataset acquisition section. Once the dataset was ready, numerous models were evaluated to ensure compatibility with the chosen hardware and to meet the stringent performance requirements. Mistaking a rock as non-dangerous could have catastrophic consequences for the robotic platform, underscoring the need for precision in model selection.

The final model chosen was DeepLabV3+ \citep{chen2018encoder}, a well-established model previously tested in rover navigation settings\citep{chiodini2020evaluation}. DeepLabV3+ offers a simple yet effective decoder module that refines segmentation results, particularly along object boundaries, enhancing obstacle definition for navigation purposes. The model takes an RGB image from the rover's front camera as input and generates an image of the same resolution, with a palette comprising five possible classes: not labelled, soil, close-rock, far-rock, and little-rock. This output is not only valuable for analysis from a human perspective, as it can be visualised in colours, but also for navigation purposes.

The palletised output is overlaid onto the depth image, where pixels corresponding to soil or little-rock are set to infinity, indicating the absence of obstacles in those directions. This effectively clears the map, enabling safer navigation. Unlabeled pixels also play a crucial role; when the rover encounters an unknown object (e.g., a person, not labelled in this dataset), the depth map is preserved in that volume, treating it as an obstacle for navigation.

One notable achievement of the semantic segmentation system was its ability to deliver results at a high frame rate, which was crucial for real-time navigation. This ensured that the rover could process and react to the environment swiftly, further contributing to the success of the mission.

In this context, it's important to acknowledge that despite these significant achievements, one of the most formidable challenges encountered during the CISRU scenario was related to communication. Operating in remote, extraterrestrial environments posed considerable difficulties in maintaining a robust and continuous data link with the rover. In this context, the palletised outputs generated by the semantic segmentation system played a critical role. Their efficiency and compact representation facilitated the transmission of crucial navigation data even under challenging communication conditions. These palletised outputs helped mitigate the impact of communication constraints, allowing the rover to continue its mission with a higher degree of autonomy and safety.

In essence, semantic segmentation, implemented through DeepLabV3+ in this scenario, serves as a cornerstone for enabling accurate and safe navigation in challenging extraterrestrial environments, where traditional depth sensors are not yet available for autonomous robotic systems.

\begin{figure}[h!]
    \centering
    \includegraphics[width=\columnwidth]{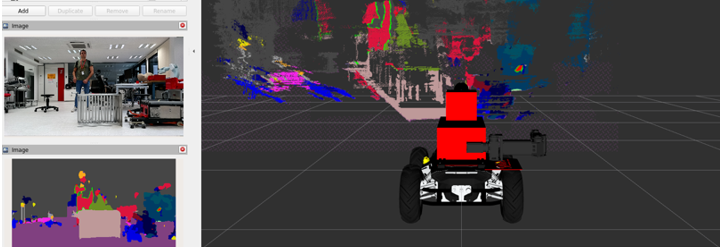}
    \caption{Semantic segmentation module}
    \label{fig:semseg}
\end{figure}

\section{AI dataset acquisition}

The use of artificial intelligence will fuel up complex robot-robot and astronaut-robot interactions in space during the next decades \citep{CISRU2022}.

One of the most critical parts of developing artificial intelligence algorithms is training. Training the AI with a quality real-image datasets rather than exclusively using synthetic images greatly increases the accuracy of the AI. 

The creation of a real-image dataset is a resource heavy process, especially in regards to the man hours required to label the images. Creating a system that would allow for self-labelling of the thousands of images constituting a dataset would save on both the time and overall monetary cost of a project.  

To carry out the developments of the second module, focusing on perception through AI, a good dataset of real or analogue data is needed to train the various models developed. With labels such as astronaut, rover and rock, not usually found in commercial or benchmark datasets.

To achieve this, three datasets of analogue data taken at the Automation and Robotics Planetary Laboratory of ESTEC have been merged using a novel self-labelling system. Additionally, data from the GMV Mars Yard SPoT and the analogue habitat in Hawaii, HI-SEAS (The Hawai'i space exploration analog and simulation), located on the slopes of Mauna Loa, have also been included.

\begin{figure}[h!]
    \centering
    \includegraphics[width=\columnwidth]{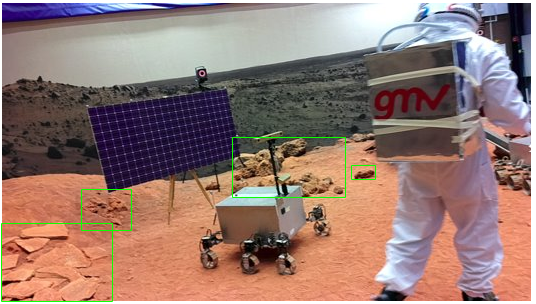}
    \caption{Rock dataset captured at the Automation and Robotics Planetary Laboratory of ESTEC}
    \label{fig:label}
\end{figure}

These datasets were labelled with bounding boxes for detection models of any kind. The labels were also exported in YOLO and PASCAL voc formats, to allow a wider range of models to test with during the project. The process of labelling was pseudo-automated: given that no previous dataset targeted this labels, a few images were hand labelled and then a Mobilenet-SSD was trained with these to label others. Then another adversarial network was added, to determine if the image was properly labelled. Additionally, at random the images were supervised to ensure the good quality of the dataset created.

This format (identification on image) of the labels proved very helpful for Human-Environment interaction detection and emergency situations, but proved to be not enough for the navigation model, in which the detection of a rock without establishing the boundaries of it didn't fix the noisy map. Due to this problem, a rock segmentation dataset from \citep{math9233048} have been used to pre-train the semantic segmentation module.

\section{Testing}

During the summer of 2023, the final tests of the project were conducted, with a focus on verifying the different capabilities of each module. To accomplish this, two distinct scenarios were prepared: one involving multi-robot planetary exploration focused on mapping and searching for materials for ISRU and the other centred around astronaut-robot collaboration focused on settlement maintenance like solar panel repair.

In the first scenario, a leading robot called the Lunar and Martian Autonomous Reconnaissance Rover (LAMARR) explores the surface, analysing the local geology in search of points of interest. Once such points are discovered, the second rover, our Mini-Autonomous Explorer (MAE), which is also mapping the terrain, approaches the first robot to provide it with the necessary tools for analysing the area. Thus, LAMARR, equipped with a robotic arm and an integrated tool-changing system, can retrieve the tools carried by MAE and also deposit regolith samples in it, allowing MAE to transport them back to the base.

The Multi-Agent was able to change the exploration plan when LAMARR was detecting a interesting zone. In this case, the plan was modified in order to obtain a sample with the cooperation of both vehicles and after finishing it, both vehicles receive the command to continue with the original exploration plan.

\begin{figure}[h!]
    \centering
    \includegraphics[width=\columnwidth]{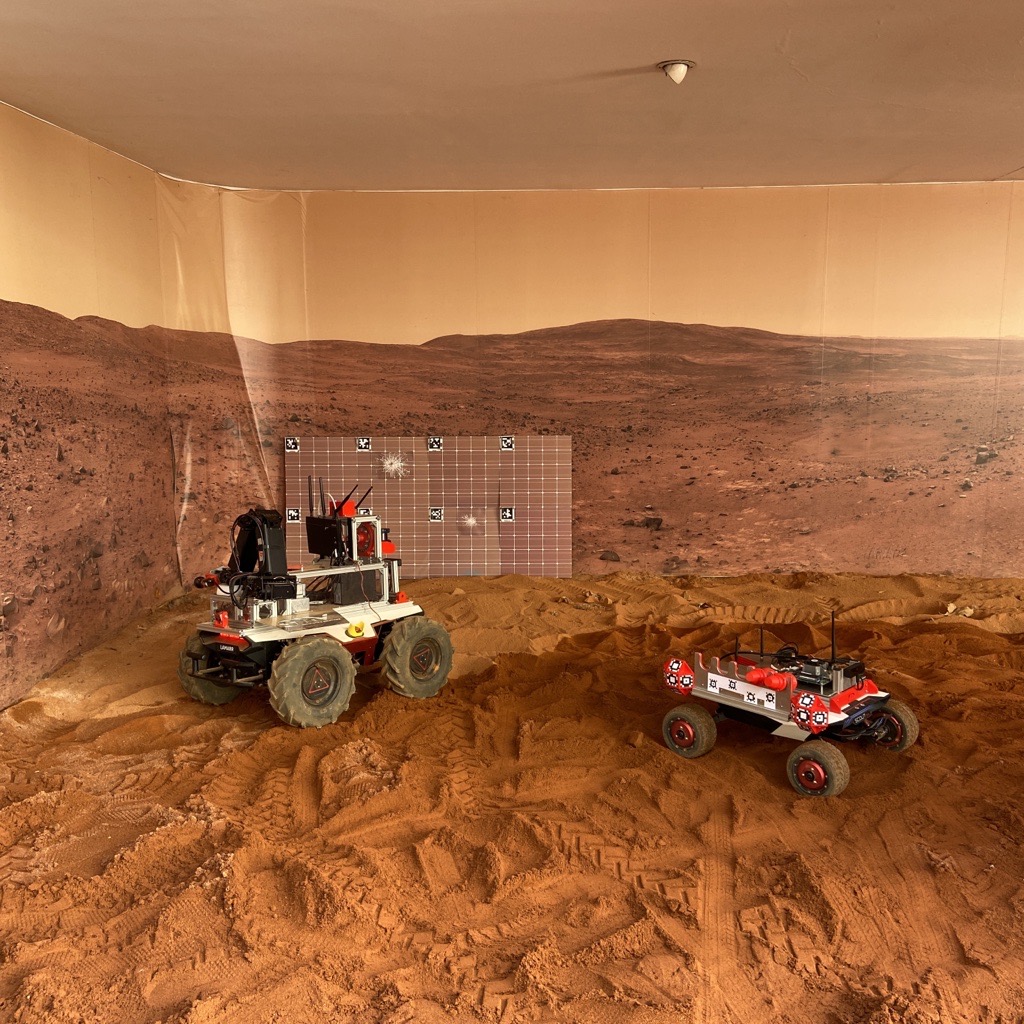}
    \caption{Lunar and Martian Autonomous Recognisance Rover (LAMARR) and Mini Autonomous Explorer (MAE) during the tests at GMV SPoT}
    \label{fig:lamarrandmae}
\end{figure}

\begin{figure}[h!]
    \centering
    \includegraphics[width=\columnwidth]{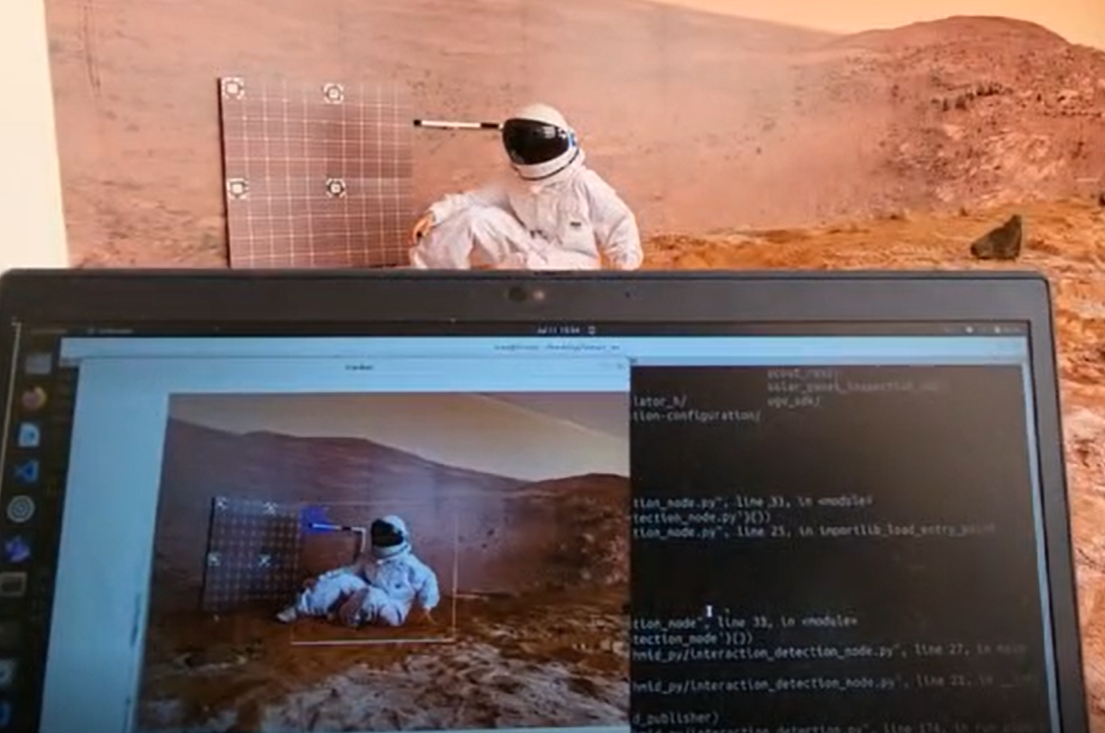}
    \caption{Astronaut drop detection}
    \label{fig:astronautemerg}
\end{figure}

The second scenario is based on activities necessary for a lunar settlement maintenance reducing the EVA and reducing the human risk. In this context, LAMARR assesses the condition of the base's solar panels and identifies any damages. If any issues are detected, the multi-agent sends collaboration requests to the astronaut, indicating which panel is damaged and describing the type of damage which is AI output. This initiates a sequence of repair tasks for the astronaut, with LAMARR providing assistance and supervision. In case of an emergency, LAMARR sends messages to the mission control centre, like if the human machine interaction detection components detects a astronaut felt as shows Fig. \ref{fig:astronautemerg}.

The semantic segmentation and astronaut interaction and emergencies detection are required for the astronaut and rover collaboration. The semantic segmentation was used not only to feed the GNC in order to avoid the noisy mapping generated by pure stereo navigation, but also to generate a safe zone around the astronaut workspace, to increase security in the mission. On the other hand, the rover was able to supervise the astronaut tasks by the human-environment detection and it was able to send notifications to the human-machine interface in case the detection of the interaction was wrong. Also, in case of astronaut drop detection, the rover was able to send an alert to the astronaut to confirm an accident. In case no answer is registered or the astronaut replies with the command HELP, the rover is able to contact the control centre to supervise the astronaut health by another person.

\section{Conclusions}

CISRU paves the way for future space exploration, where humans and robots must collaborate to survive and create new infrastructure. Moreover, CISRU also serves as a transfer system between space technology and terrestrial applications, with potential use cases in the nuclear sector, refineries, mining, and more.

The development of analog datasets is a key driver in the utilisation of AI in space, as well as its integration into space-graded systems and compliance with space software requirements. One of the major challenges with these datasets is labeling, which we have addressed by creating a new method that combines different data sources and enables automated labeling. This significantly reduces costs in terms of manpower.

It is worth noting that an interdisciplinary approach is essential for these developments, enabling us to understand the diverse mission needs and create an abstract and user-friendly suite.

\section*{Acknowledgments}

This project is a testament to the strong collaboration between GMV and the European Space Agency (ESA) in achieving lunar exploration objectives. We must express our gratitude to ESA for their trust in us in developing this software suite, which will serve as a starting point for collaborative and inclusive robotics beyond Earth.

\printbibliography

\end{document}